\documentclass{article}

\usepackage{url}
\usepackage[hidelinks]{hyperref}
\usepackage[utf8]{inputenc}
\usepackage[small]{caption}
\usepackage{graphicx}
\usepackage{amsmath}
\usepackage{amssymb}
\usepackage{amsthm}
\usepackage{booktabs}
\usepackage{algorithm}
\usepackage{algorithmic}
\usepackage{verbatim}
\usepackage{microtype}
\usepackage{tabularray}
\usepackage{makecell}
\usepackage{listings}
\usepackage{authblk}
\usepackage{helvet}
\usepackage{breakurl}
\usepackage{tikz}
\usepackage{pgfplots}
\pgfplotsset{compat=newest}
\definecolor{taupurple}{HTML}{4E008E}
\definecolor{taured}{HTML}{F07387}
\definecolor{taupink}{HTML}{F5A5C8}
\definecolor{tauyellow}{HTML}{FFDCA5}
\definecolor{tauwhite}{HTML}{FFFFFF}
\definecolor{taulpurple}{HTML}{C3B9D7}
\definecolor{taulblue}{HTML}{82C8F0}
\definecolor{taulgreen}{HTML}{7DCDBE}
\definecolor{taulgrey}{HTML}{C8C8C8}

\newcommand*{\PL}{\mathrm{PL}}
\newcommand{\union}{\cup}
\newcommand{\isect}{\cap}
\newcommand{\set}[1]{\{#1\}}
\newcommand{\pair}[2]{\langle{#1},{#2}\rangle}

\newcommand{\sel}[2]{\set{{#1}\mid{#2}}}
\newcommand{\rg}[3]{{#1}{#2}\ldots{#2}{#3}}
\newcommand{\eset}[2]{\set{\rg{#1}{,}{#2}}}
\newcommand{\lequiv}{\leftrightarrow}
\newcommand{\at}[1]{\tau_{#1}}
\newcommand{\datom}[3]{p(#1,#2,#3)}
\newcommand{\lm}[1]{\mathrm{LM}(#1)}
\newcommand{\GLred}[2]{{#1}^{#2}}
\newcommand{\IF}{\leftarrow}

\newcommand{\code}[1]{\texttt{#1}}
\newcommand*{\eofex}{\mbox{}\nobreak\hfill\hspace{0.5em}$\blacksquare$}

\newtheorem{example}{Example}

\title{Globally Interpretable Classifiers via Boolean Formulas with Dynamic Propositions}

\author{Reijo Jaakkola}
\author{Tomi Janhunen}
\author{Antti Kuusisto} 
\author{Masood~Feyzbakhsh~Rankooh}
\author{Miikka Vilander\footnote{The authors are given in the alphabetical order.}}
\affil{Tampere University}

\date{}

\begin{document}

\maketitle

\begin{abstract} 
\noindent Interpretability and explainability are among the most important challenges of modern artificial intelligence, being mentioned even in various legislative sources. In this article, we develop a method for extracting immediately human interpretable classifiers from tabular data. The classifiers are given in the form of short Boolean formulas built with propositions that can either be directly extracted from categorical attributes or dynamically computed from numeric ones. Our method is implemented using Answer Set Programming. We investigate seven datasets and compare our results to ones obtainable by state-of-the-art classifiers for tabular data, namely, XGBoost and random forests. Over all datasets, the accuracies obtainable by our method are similar to the reference methods. The advantage of our classifiers in all cases is that they are very short and immediately human intelligible as opposed to the black-box nature of the reference methods.
\end{abstract}

\section{Introduction}

Human interpretability is one of the main challenges of modern artificial  intelligence (AI). This has lead to an increasing interest in explainable AI (or XAI for short). 
The right for receiving explanations is already mentioned even in various legislative sources such as the European Data Protection Regulation \cite{euregulationstuff} and the California Consumer Privacy Act \cite{california-oag}. 

Interpretability is typically divided into \emph{local} and \emph{global} variants. The former relates to providing reasons why a classifier produced a particular decision with a particular input, while global interpretability means that the entire behaviour of the classifier---on any input---can be understood due to a single and sufficiently clear explication.

The current paper introduces and investigates a method for producing globally interpretable classifiers for binary classification tasks on tabular data. The method is implemented declaratively by deploying the paradigm of Answer Set Programming (ASP; see \cite{BET11:jacm} for an overview) and, more specifically its theory-based extension ASP \emph{modulo difference logic}, abbreviated ASP(DL) \cite{JKOSWS17:tplp}. Using simple difference constraints, the discretization of numeric values takes place dynamically during the search for actual (optimal) explanations. As shown in this work, Boolean statements resulting from such dynamic discretization offer more fine-grained primitives for classifying data points and give rise to even shorter classifying formulas compared to an earlier static approach based on medians \cite{jelia23}.

The global nature of the interpretability stems from the fact that the formulas acting as classifiers are very short Boolean assertions about the attributes of the original datasets. Thus, the formulas read as very simple statements in natural language about the data studied. 
E.g., the formula
\[ \neg (p_{\geq 6} \lor q_{\geq 7} \lor r_{\geq 5}) \]
is an explanation that we obtain from our experiments on a dataset concerning breast cancer. The predicate $p_{\geq 6}$ states that the value of $p$, a measure relating to bare nuclei, is at least $6$. The interpretations are similar for the other predicates, with $q$ and $r$ denoting measures of clump thickness and uniformity of cell size, respectively. The formula is clearly short and it is immediately clear (at least to an expert from the respective medical field) what the general meaning of the formula is. The accuracy of this formula on the holdout data is \(97.1\) percent.

While this paper is focused on obtaining explanations from data, we note that our method could also be used to explain black-box classifiers. In this case, one would first use the black-box classifier to generate data and then explain that data as described in this paper.

\subsection{The formula-size method} 

We next give an overview of our method; for the 
full details, please see Section \ref{sec:expmethod}. 
Let $M$ denote a tabular dataset 
with the set $\{Y_1, \dots , Y_{s}\} \cup 
\{Z_1, \dots , Z_{k}\}$ of attributes, each $Y_i$ being 
numeric and $Z_i$ categorical. To enable Boolean representations of
classifiers, we model each $Y_i$ and $Z_j$ by Boolean propositions
defined as follows. 
\begin{enumerate} 
\item 
For each number $r$ in the range of 
the attribute $Y_i$ in the dataset $M$, we define a corresponding
proposition $p_i^{\geq r}$.
\item 
Applying a natural one-hot encoding, we 
define, for each category $C$ of $Z_j$, a
corresponding proposition~$p_{j}^C$. 
\end{enumerate}

These propositions are interpreted in $M$ in a natural 
way, meaning that a data point $w$ satisfies $p_{i}^{\geq r}$ 
(respectively, $p_{j}^{C}$) if $w$ has the value of $Y_i$ at least $r$ (respectively, belongs to the category $C$ of $Z_j$). 

Given this typically huge set $\Pi$ of 
proposition symbols at our 
disposal, we do
the following. We randomly split the data into 
two parts: a 70 percent part $W_{70}$ for training and a 30 percent part $W_{30}$ for validation. We consider the accuracies of formulas over the training and validation sets. By accuracy we mean the percentage of points on which the formula agrees with the target attribute of the binary classification task.
In the training part, we do the following for
increasing \emph{length bounds} $\ell$, starting with $\ell = 1$. 

\begin{enumerate} 
\item 
Among all Boolean formulas of length at most $\ell$ and
using the propositions in $\Pi$, we
find a formula $\varphi_{\ell}$ with  
maximum accuracy in $W_{70}$. 
\item 
We record the 
accuracy $\mathit{acc}_{30}(\varphi_{\ell})$ of $\varphi_{\ell}$ 
in the validation set $W_{30}$. 
We then repeat the above step $1$ with $\ell := \ell + 1$.\end{enumerate} 
Using early stopping, we scan for longer and 
longer formulas $\varphi_{\ell}$ until overfitting. Our early stopping trigger checks when
$\mathit{acc}_{30}(\varphi_{\ell})$ is smaller than the best accuracy obtained so far, twice in a row. Supposing $\ell$ is the greatest value we reach before overfitting, we then compute the final classifier formula $\psi$ by using the the formula bound $\ell$ and the union $W$ of the training and valuation data. 

We also test two other variants of the formula-size method. A more restricted version we introduced in \cite{jelia23} uses a system based on propositions $p_{\geq  r'}$ where $r'$ is always the median of the corresponding numeric attribute. We discuss this further below in the section devoted for related work. Another more general version has proposition symbols that specify an interval $[r,r']$ instead of a single pivot value $r$. All other details are the same for our experiments on these variants of the method and we compare all three in terms of formula size as well as accuracy.

To test our method, we use a ten-fold cross validation procedure.
This means that each dataset $W$ in the actual experiments is 90 percent of the very original dataset studied, with 10 percent reserved as holdout data. 

Our method is implemented in ASP \cite{BET11:jacm} which brings along the strengths of declarative programming: the logic programs can be understood as \emph{logical specifications} of the concepts involved such as formulas forming the hypothesis space, the objective function used to measure their quality, and the schemes used to discretize the data provided as input. Since logic programs are highly elaboration tolerant, it is easy to formalize these concepts one-by-one in an incremental way.
Moreover, we advocate the overall modularity of the implementation by the three-layer architecture illustrated in Figure~\ref{fig:layers}. As a consequence, it is straightforward to experiment with different encodings as well as alternative concepts.

It is worth noting that the number of Boolean formulas grows asymptotically extremely fast, with the number of non-equivalent formulas of size at most \(k\) being exponential in relation to both \(k\) and the number of propositions used. While symmetry reductions and other techniques can be deployed to reduce the hypothesis space significantly, it is still natural to question how the method described above could be feasible in the first place.
The answer relates to overfitting. It is clear that if too long formulas are allowed, they will overfit. Therefore, we need to use early stopping to avoid overfitting to the training data. In a large majority of our experiments on real-life datasets, signs of overfitting emerge very fast, triggering early stopping and leading to short formulas and feasible runtimes. Furthermore, such short formulas yield accuracies similar to reference methods. The fact that short formulas suffice indicates that, in a sense, asymptotic complexity is not the right measure here. This is 
important to our method, since too long formulas would not be inherently interpretable anyway.

\subsection{Results}

We run our method on seven datasets originating from the UCI machine learning repository. 
As tabular data is still
challenging to deep learning approaches \cite{grinsztajn2022why,tabularsurvey}, we use XGBoost \cite{xgboost} and Sklearns implementation of random forests \cite{breiman2001random} as our comparison methods. These methods are 
esteemed state-of-the-art approaches for tabular data \cite{swartzarmon22,grinsztajn2022why} and thus suitable for comparisons. 

As mentioned above, we compare three different variants of our method: the static median-based formula-size method of \cite{jelia23}, a dynamic discretization based on single pivot points $r$ and a version with intervals $[r,r']$.
Here the focus is seeing how much improvement the different approaches to dynamic discretization can provide. 
From Figure \ref{fig:resultsum1} and Tables \ref{tab:interval_formulas}, \ref{tab:pivot_formulas} and \ref{tab:median_formulas}, one can see that for several datasets, the dynamic approaches either improve the accuracy or decrease the formula size needed to obtain a similar accuracy.

The average accuracy and standard deviation of each method over ten-fold cross-validation are reported in Figure \ref{fig:resultsum1}. For all datasets, we obtain immediately interpretable formulas with accuracy similar to the reference methods. Formulas obtained as classifiers are reported in Table~\ref{tab:interval_formulas} for the interval method, in Table~\ref{tab:pivot_formulas} for the pivot method and in Table~\ref{tab:median_formulas} for the median method. The tables also list average formula lengths over all runs. The example formulas in the tables were taken from the first round of the ten fold cross validation in each case. 

As an example result, consider the classifier 
\[p_{[25,60]} \land q_{[128,196]} \]
obtained with the interval method from the Diabetes dataset. Here the attribute $p$ is the age of the person and the attribute $q$ is their measured glucose value. The formula offers high glucose and age between 25 and 60 as an explanation for diabetes.
Another example is the formula 
\[
p \lor q_{\geq 767}
\]
obtained with the pivot method from four different splits of the BankMarketing dataset (with slight variations on the bound of $q$). The classification task concerns whether a marketing call resulted in subscription for a term deposit. Here the attribute $p$ indicates whether the customer has previously subscribed for a deposit and the attribute $q$ gives the duration of the call in seconds. So a customer is likely to subscribe to a deposit if they have done so previously or the telemarketer can keep them on the phone for a long duration. The accuracies of this formula on the holdout data ranged from \(88.1\) percent to \(90.3\) percent for different splits.

We conclude that in all cases, our classifiers are immediately globally interpretable and   the accuracy is similar to the ones obtained by XGBoost and random forests. 

The interpretability is a real gain, as the classifiers produced by XGBoost and random forests are of a black-box nature. Indeed, the classifiers returned by these methods are ensembles of decision trees, with at least tens or even thousands of members.

A shortcoming of our method is that it is computationally demanding and thus not feasible in scenarios where classifiers must be obtained fast. In the worst case, the experiments with our method took days in a high-power computation environment, while the comparison methods could be completed much faster with an ordinary laptop. 
The runtimes are discussed in Section 
\ref{sec:runtimes}. 

\subsection{Related work}

Explainable AI is a large and active field, with 
logic-based explainability being a 
growing subfield.
See \cite{silva} for a thorough survey on logic-based explainability.
Much of the recent work on this topic has been initiated
 in \cite{ShihCD18} and \cite{alexey}, where natural 
minimality notions such as prime
implicants are utilized as explanations.

Much of the work on logic-based explainability focuses on local aspects of explainability. The current paper, on the other hand, focuses on global explainability using short formulas of Boolean logic. It is generally agreed that small classifiers
can be regarded as global explanations
\cite{OptimalRuleLists,BertsimasD17}. In this work we aim for very small formulas, following a ``simplicity first'' philosophy.
Relating to this ideology, already \cite{verysimplerules1993} reported that in many of the 16 datasets studied there, even a single attribute was sometimes enough to get an accuracy not too drastically smaller than one obtainable by the then state-of-the-art decision trees. 

Other approaches to global explainability using small classifiers include the articles \cite{BertsimasD17,OptimalRuleLists}. These studies investigate the use of small rule lists and decision trees, which are optimal with respect to size and training error. The empirical results reported in these papers also demonstrate the surprising effectiveness of interpretable models on real-world tabular data. For further related work on global explanations and interpretable AI, see the survey \cite{rudin2019explaining}. 

The quest for short explanations is compatible with the principle of Occam's razor \cite{BEHW87:ipl} and the early search-based approaches, such as current best hypothesis search \cite{Mitchell82:aij}, already implement this idea. However, the performance of computers has improved drastically since then and the solver technology used to implement the search for hypotheses in the present work has emerged by the side.

In this work, we continue the line of research initiated in \cite{jelia23} where numeric attributes are still made Boolean statically, with the intuitive reading ``the value of the attribute is above median'' for Boolean variables. That paper presents three experiments with single split cross validation, obtaining related Boolean  classifiers.
In the current paper, we redo the experiments of \cite{jelia23} with the seven datasets studied, two of which were used in \cite{jelia23}. As mentioned above, the related results are in Figure \ref{fig:resultsum1} and Tables~\ref{tab:interval_formulas}, \ref{tab:pivot_formulas} and \ref{tab:median_formulas}.
While reasonable classifiers are 
obtained, the ones using dynamic predicates $p_{[r,r']}$
and $p_{\geq  r}$ lead to formulas that are shorter or more
accurate, while keeping the running times feasible.

\section{Preliminaries}
\label{sec:preliminaries}

A finite set $\tau = \{p_1, \dots, p_k\}$ of proposition symbols is called a \textbf{Boolean vocabulary}. The syntax of propositional logic over the vocabulary $\tau$, denoted $\PL[\tau]$, is given by the following grammar:
\[
\varphi ::= p \mid \neg \varphi \mid \varphi \lor \varphi \mid \varphi \land \varphi
\]
where $p \in \tau$. 
We also define the equivalence connective as the abbreviation $\varphi \leftrightarrow \psi := (\varphi \land \psi) \lor (\neg \varphi \land \neg \psi)$.

A \textbf{Boolean $\tau$-model} is a structure $M = (W, V)$, where the set $W \neq \emptyset$ is the \textbf{domain} of $M$ and the function $V : \tau \to \mathcal{P}(W)$ is a \textbf{$\tau$-valuation}. For $p \in \tau$, the set $V(p) \subseteq W$ is the set of points, where the proposition $p$ is considered to be true.
Given a $\tau$-valuation $V$, we extend it in the standard way to a valuation $V : \PL[\tau] \to \mathcal{P}(W)$ concerning all $\PL[\tau]$-formulas. 

The \textbf{size} of a formula $\varphi \in \PL[\tau]$ is defined as the number of occurrences of proposition symbols and connectives in the formula. For example the size of \(\neg (p \land r)\) is 4.

Let $M = (W, V)$ be a $\tau \cup \{q\}$-model and let $\varphi$ be a $\tau$-formula. The \textbf{accuracy} of $\varphi$ over $M$ is 
\begin{equation}\label{eq:accuracy}
    \mathrm{acc}(\varphi, M) := \frac{|V(\varphi \leftrightarrow q)|}{|W|}.
\end{equation}
The accuracy of $\varphi$ is the percentage of points where $\varphi$ and the target attribute $q$ agree. 

We move on from propositional structures to structures with both Boolean and numeric attributes. The datasets we consider also have categorical attributes with more than two categories, but we use one-hot encoding to Booleanize the categorical attributes as a preprocessing step. See Subsection \ref{sec:experimentsetup} for more discussion on this.

A \textbf{Boolean attribute} over a set of data points $W$ is simply 
a subset of $W$. As seen above, the interpretations of proposition symbols $p$ in propositional logic are Boolean attributes $V(p) \subseteq W$. 
A \textbf{numeric attribute} over $W$ and with a finite codomain $R$ is a
function $f : W\rightarrow R$. We denote the set of functions $f: W \to R$ by $R^W$. In this paper, $R$ is a set of floating point numbers and thus always finite. We assume implicitly that the codomain $R$ has a linear order $\leq$.

A \textbf{general vocabulary} $\mathcal{X} = \mathcal{B} \uplus \mathcal{N}$ is a
disjoint union of a 
Boolean vocabulary $\mathcal{B} = \{p_1, \dots , p_{\ell}\}$ and a 
\textbf{numeric vocabulary} $\mathcal{N} = \{s_1, \dots , s_m\}$. A 
\textbf{general valuation} $V$ over $\mathcal{X}$ and $R$ is a
function $V: \mathcal{X} \rightarrow \mathcal{P}(W) \cup R^W$ which splits naturally into 
two functions $V_B : \mathcal{B} \rightarrow \mathcal{P}(W)$ and
$V_N : \mathcal{N} \rightarrow R^W$. 
For convenience, we write $s_i^V$ instead of $V(s_i)$. 
A \textbf{general structure} is a tuple $(W, V)$, where $W$ is a set of data points and $V$ is a general valuation. 

We define two different ways to discretize general vocabularies and structures based on a single pivot value $r \in R$ and an interval $[r,r'] \subseteq R$, respectively. We first define the pivot discretizations.

Let $\mathcal{X} = \mathcal{B} \uplus \mathcal{N}$ be a general vocabulary. 
The \textbf{complete pivot vocabulary} over $\mathcal{N} = \{s_1, \dots, s_m\}$ and $R$ is the (finite) set of proposition symbols $p_{s_i}^{\geq r}$
where $s_i \in \mathcal{N}$ and $r \in R$. Intuitively, the complete dynamic vocabulary contains all possible propositions $p_{s_i}^{\geq r}$ one might need when Booleanizing the numeric attributes of $\mathcal{N}$ using a single pivot point. 

A Boolean vocabulary $\eta$ is a \textbf{$R$-pivot discretization} of a numeric vocabulary $\mathcal{N}$ if $\eta$ is the range of  an
injection $f$ from $\mathcal{N}$ into the 
complete dynamic vocabulary over $\mathcal{N}$ and $R$ such that for all $s \in \mathcal{N}$, $f(s) =  p_{s}^{\geq r}$ for some $r\in R$. Following the intuition of discretization, this 
turns every numeric attribute symbol $s$ into a Boolean 
symbol $p_{s}^{\geq r}$. 

Let $M = (W, V)$ be a general structure over $\mathcal{X} = \mathcal{B} \uplus \mathcal{N}$. 
A Boolean structure $M' = (W, V')$ is a \textbf{pivot discretization} of $M$
if the following conditions hold.
\begin{enumerate} 
\item 
$V'$ is a function $V' : \mathcal{B} \uplus \eta \to \mathcal{P}(W)$,
where $\eta$ is a $R$-discretization of $\mathcal{N}$. 
\item 
$V'(p) = V(p)$ for every $p \in \mathcal{B}$. 
\item 
$V'(p_{s}^{\geq r}) = \{w \in W \mid s^V(w) \geq  r\}$ for all $p_{s}^{\geq r} \in \eta$.
Recall that $s^V$ denotes the function $V(s):W\rightarrow R$ that we obtain from $V$ when given the input $s$. 
\end{enumerate} 

The \textbf{pivot class} over $M$ is the set of all pivot discretizations of $M$,  denoted by $\mathrm{PD}(M)$.

We similarly define the notions of \textbf{complete interval vocabulary}, \textbf{$R$-interval discretization}, \textbf{interval discretization} of $M$ and \textbf{interval class} $\mathrm{ID}(M)$ over $M$ by simply replacing the condition $\geq r$ with inclusion in an interval $[r,r']$.

Let $\mathcal{F}$ be a set of Boolean formulas and 
$\mathcal{M}$ a set of Boolean models (possibly over 
different vocabularies). The \textbf{maximum 
accuracy} over $\mathcal{F}$ and $\mathcal{M}$ is the maximum 
number $\mathrm{acc}(\varphi, M)$ such that 
\begin{enumerate} 
\item 
$\varphi \in \mathcal{F}$ and $M \in \mathcal{M}$,
\item 
only propositions from the vocabulary of $M$ occur in $\varphi$.
\end{enumerate} 
The Boolean formula $\varphi$ is then said to \textbf{realize} the 
maximum accuracy (over the model $M$). 

Let $M$ be a general
structure over $\mathcal{X} = \mathcal{B} \uplus \mathcal{N}$ and $R$. 
A Boolean formula $\varphi$ is a \textbf{$M$-pivot-formula} (resp. \textbf{$M$-interval formula}), if $\varphi$ is a Boolean formula over a Boolean vocabulary $\mathcal{B} \uplus \eta$, where $\eta$ is a $R$-pivot discretization (resp. $R$-interval discretization) of $\mathcal{N}$. We denote the set of all $M$-pivot-formulas by $\mathrm{PF}(M)$ and the set of all $M$-interval-formulas by $\mathrm{IF}(M)$.

For $\ell \in \mathbb{N}$, we denote the set of $M$-pivot-formulas with size at most $\ell$ by $\mathrm{PF}_\ell(M)$. We define $\mathrm{IF}_\ell(M)$ in the same way.

When discussing real-life datasets in the other sections below, we will suppress the model notation $M = (W, V)$ and only discuss the set $W$ of data points as well as some subsets of $W$, to be interpreted as submodels of $M$.

\section{The dynamic formula length method}\label{sec:expmethod}

We next give a step-by-step description of our method. We describe the method for pivot discretization; the case of interval discretization is identical with pivot-formulas switched for interval-formulas. The median method uses a static discretization based on the median; the other details are the same as with the other variants.

Note that this procedure assumes we have already 
separated ten percent of the original dataset to be 
used as a final testing data. Thus the input to 
the process is a dataset $W$ consisting of the
remaining 90 percent. 

In the procedure, we scan through all formulas 
up to increasing formula length
bounds. We shall describe in 
Section \ref{sec:implementation} how this can be done reasonably 
efficiently, avoiding a naive brute force search. 
We use early stopping to avoid overfitting and formulas too long to be interpretable.

The method consists of the following stages. 

\begin{enumerate} 
\item 
An input to the method is a tabular 
dataset $W$ with binary and numeric attributes $X_1, \dots , X_{m},q$,
where $q$ is a binary target attribute. 
From this data $W$, we 
randomly separate a 30 
percent set $W_v\subseteq W$ to be used as 
\emph{validation data}. The remaining 70 
percent $W_{\mathit{tr}}\subseteq W$ is called the 
\emph{training data}. 
\item 
For a length bound $\ell$ starting with 1, we scan through the set $\mathrm{PF}_\ell(W_{\mathit{tr}})$ of all $W_{\mathit{tr}}$-pivot-formulas $\varphi$ 
with maximum length $\ell$, choosing the formula
$\varphi_{\ell}$ that realizes the maximum accuracy over the training
set $W_{\mathit{tr}}$. 
\item 
For each chosen formula $\varphi_{\ell}$, we record its accuracy over the validation set $W_v$. Then, we update the best validation accuracy so far, denoted $\Delta$.
Now, if the accuracies of both $\varphi_{\ell-1}$ and $\varphi_{\ell}$ are smaller than $\Delta$, we trigger early stopping and move on to step 4. Otherwise, we return to step 2 with $\ell$ increased by one.
\item
We scan through the formulas 
\[\varphi_1, \dots , \varphi_{\ell-1}\]
again and find the smallest
number $L\in \{1, \dots , \ell-1\}$ such that
the accuracy of $\varphi_L$ over $W_{v}$ is the same as that of $\varphi_{\ell-1}$. 
This step ensures that if many 
formulas of different lengths obtain the same accuracy, we
choose the smallest length, leading to shorter final formulas.
\item 
Finally, we use the data $W = W_{tr}\cup W_{v}$ and the length bound $L$ to compute the final output formula by 
scanning through the set $\mathrm{PF}_L(W)$ of all $W$-pivot-formulas $\psi$ 
with maximum length $L$, and then choosing the formula
with the maximum accuracy over $W$. 
\end{enumerate}  

We emphasize that the search conducted in steps 2 and 5 for the optimal pivot-formula is not carried out as a brute force search. Even though the search remains a challenging problem, we use symmetry reductions and other techniques to reduce the size of the hypothesis space to make it practically feasible to conduct the search on real-life datasets. See Section \ref{sec:hypotheses} for more details. Our results in Section \ref{sec:results}, along with the training times we report in Section~\ref{sec:runtimes}, show that the method is useful in practice.

In step 3, we stop once the accuracy of the formulas on the 30-percent validation data is worse than the best accuracy so far, twice in a row. This early stopping is done to avoid overfitting. In general, as the length bound of the formulas increases, the accuracy on both the training data $W_{\mathit{tr}}$ and the validation data $W_v$ improves at the start. At some point, however, the validation accuracy will start to decline while the training accuracy keeps on improving. This is when the formulas start to overfit to the training data. Our simple early stopping trigger aims to avoid overfitting and also obtain short, interpretable formulas. 

Finally in step 5 we use the formula length found in step 4 to compute the final classifier using all of the available data. This is to ensure that our early stopping procedure does not lead to throwing away 30 percent of the data.

Note that even though our method gives a way to select a unique classifier, it can also be worthwhile to consider the full sequence of formulas with increasing lengths. This sequence gives an evolving picture of explanations that can be found from the data. 

\section{Implementation}\label{sec:implementation}

In this section, we present an implementation of dynamic propositions
in ASP and, in particular, its theory-style extension by simple linear
constraints known as
\emph{difference constraints}.
The respective extension of ASP is known as
ASP \emph{modulo difference logic} (ASP(DL)) \cite{JKOSWS17:tplp}
and it is natively implemented by the \emph{Clingo-dl} system\footnote{\url{https://github.com/potassco/clingo-dl}}.
Difference constraints have a normal form $x-y\leq k$ where $x$ and
$y$ are variables and $k$ is a constant. For the purposes of this
work, the domain of integers is sufficient, since the relative orders
attribute values are significant rather than their magnitudes.  It is
also worth noting that satisfiability can be checked very efficiently
for sets of difference constraints, e.g., by using the Bellman-Ford
algorithm \cite{Bellman58:appmath}.

Without going into syntactic details of rules used in logic programs,
we recall that the \emph{answer sets} of a logic program $P$ with a
vocabulary $\at{P}$ are sets of atoms $S\subseteq\at{P}$ satisfying a
fix-point equation $S=\lm{\GLred{P}{S}}$ where
(i)
$\lm{\cdot}$ denotes the \emph{least model} of the positive
program given as argument and
(ii)
the reduct $\GLred{P}{S}$ is obtained by partially evaluating the
negative conditions of $P$ with respect to $S$ \cite{GL88:iclp}.
In view of definitions in Section
\ref{sec:preliminaries}, the answer sets $\rg{S_1}{,}{S_n}$ of $P$
induce a $\tau_P$-model
$\pair{\eset{1}{n}}{V_P}$
such that $V_P(a)=\sel{i}{a\in S_i}$ for $a\in\at{P}$.
If difference constraints $(x-y\leq k)$ are incorporated into a logic
program $P$, they are treated as special \emph{difference atoms}
$\datom{x}{y}{k}$ in $\at{P}$ forming the respective subsignature
$\at{d}\subseteq\at{P}$ of the program $P$.
If $\at{d}\neq\emptyset$, the set of linear inequalities
\begin{equation}\label{eq:strict-theory}
\sel{x-y\leq k}{\datom{x}{y}{k}\in\at{d}\isect S}
~\union~
\sel{x-y>k}{\datom{x}{y}{k}\in\at{d}\setminus S}
\end{equation}
must be additionally satisfied by $S=\lm{\GLred{P}{S}}$. Note that the
negations $x-y>k$ in \eqref{eq:strict-theory} can be rewritten in the
normal form as $y-x\leq -k-1$, i.e., they can be treated efficiently
as difference constraints. In contrast with traditional
\emph{satisfiability modulo theories} (SMT) approaches
\cite{BT18:handbook},
the equation \eqref{eq:strict-theory} gives a \emph{strict}
interpretation to difference atoms \cite{JKOSWS17:tplp}, since
auxiliary atoms $\datom{x}{y}{k}$ falsified by $S$ contribute the
negation $(x-y>k)$ to the theory being satisfied. This is in line with
the minimality of answer sets in general and \emph{program completion}
\cite{Clark78}: the satisfaction of \eqref{eq:strict-theory} amounts to
satisfying the completing formula $\datom{x}{y}{k}\lequiv(x-y\leq k)$
for $\datom{x}{y}{k}\in\at{d}$. When an answer set $S$ is reported to
the user, some concrete values can be derived for variables $x$ and $y$ involved in atoms $\datom{x}{y}{k}\in S\isect\at{d}$.

\begin{example}
One of the datasets studied in this paper has age as an attribute, the actual ages ranging from 21 to 81 years. Using an integer variable $a$ for an age limit, we can define its range in terms of two ASP(DL) facts:
\begin{center}
$z-a\leq -21$.~~~~$a-z\leq 81$.
\end{center}
where $z=0$ is an auxiliary variable (cf. footnote \ref{fn:missing-zero}). Assuming a particular data point where the age happens to be $27$, its Booleanization $b$ can be captured with a rule
\begin{center}
$b \IF a-z\leq 27$.
\end{center}
Using auxiliary atoms introduced above, the facts and rules above give rise to the following answer sets:
\begin{center}
$\begin{array}{rcl}
S_1 & = & \set{\datom{z}{a}{-21},\datom{a}{z}{81},\datom{a}{z}{27},b} \\
S_2 & = & \set{\datom{z}{a}{-21},\datom{a}{z}{81}}.
\end{array}$
\end{center}
The former can be backed up by assignments $z=0$ and $a=21$, while
the latter by $z=0$ and $a=28$. These minimal non-negative
values are given by \emph{Clingo-dl}, too.
\eofex
\end{example}

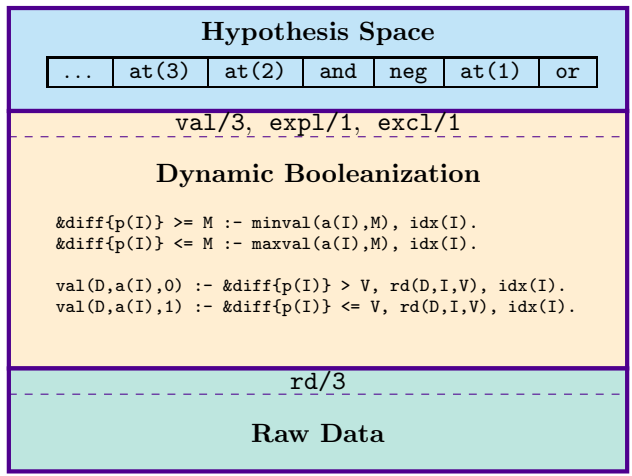
\begin{figure}[t]
\begin{center}
\begin{tikzpicture}[scale=0.685]
\draw [ultra thick,taupurple,fill=taulblue!50] (0.0,7.0) rectangle (12.0,9.0);
\node at (6.0,8.5) {\textbf{Hypothesis Space}};
\node at (6.0,7.75) {\begin{small}
\begin{tabular}{|c|c|c|c|c|c|c|}
\hline
\ldots
& \code{at(3)}
& \code{at(2)}
& \code{and}
& \code{neg}
& \code{at(1)}
& \code{or} \\
\hline
\end{tabular}
\end{small}};
\draw [ultra thick,taupurple,fill=tauyellow!50] (0.0,2.0) rectangle (12.0,7.0);
\node at (6.0,6.75) {\code{val/3},~~\code{expl/1},~~\code{excl/1}};
\draw [style=dashed,taupurple] (0.0,6.5) -- (12.0,6.5);
\node at (6.0,5.75) {\textbf{Dynamic Booleanization}};
\node at (6.0,4.0) {\begin{minipage}{0.575\linewidth}
\begin{scriptsize}
\begin{verbatim}
&diff{p(I)} >= M :- minval(a(I),M), idx(I).
&diff{p(I)} <= M :- maxval(a(I),M), idx(I).

val(D,a(I),0) :- &diff{p(I)} > V, rd(D,I,V), idx(I).
val(D,a(I),1) :- &diff{p(I)} <= V, rd(D,I,V), idx(I).
\end{verbatim}
\end{scriptsize}
\end{minipage}};
\draw [ultra thick,taupurple,fill=taulgreen!50] (0.0,0.0) rectangle (12.0,2.0);
\node at (6.0,1.75) {\code{rd/3}};
\draw [style=dashed,taupurple] (0.0,1.5) -- (12.0,1.5);
\node at (6.0,0.75) {\textbf{Raw Data}};
\end{tikzpicture}
\end{center}
\caption{Three Layers of the Implementation\label{fig:layers}}
\end{figure}

Our implementation is based on a modular architecture divided into
three layers as illustrated in Figure \ref{fig:layers}.
The topmost layer is responsible for generating hypotheses, evaluating
them with respect to data points, and calculating the overall accuracy as
part of the objective function. The middle layer encodes the
principles for Booleanizing the data and decides which attribute is
the target of explanation, and which attributes may contribute to the
actual explanations. The bottom layer extracts the values of the
attributes from the real data as a preprocessing step.
Although the architecture in Figure \ref{fig:layers}
looks \emph{stratified} at the first glance, it does not have a unique
answer set as typical for stratified programs. In particular, the two
highest layers involve choices that create a search space for
optimization.  The layers are further detailed in the respective
Sections \ref{sec:hypotheses}--\ref{sec:raw-data} below. By changing logic programs associated with the two top layers, it is also easy to take different hypothesis spaces into consideration and to change the strategy for Booleanizing attributes.

\subsection{Generating the hypothesis space}
\label{sec:hypotheses}

The topmost layer has access to the underlying data set in Boolean
form only. The data is abstracted using a 3-argument
predicate \code{val(D,A,V)} where \code{D} is the identifier of the
data point, \code{A} the index of the attribute, and
\code{V} the value, i.e., either \code{0} or \code{1}.
Besides this view to the data, the hypothesis generator receives a
target attribute any denied attributes expressed using
predicates \code{expl/1} and \code{excl/1}, respectively. Based on
this interface, the generator is completely unaware whether
the \emph{original} data is Boolean or not and its
implementation can be kept independent of this aspect.

In this research, we use the generator from \cite{jelia23} revised to
respect the interface just described. The generator represents
formulas as sequences of attributes and logical operators following
the idea of \emph{reverse Polish notation}. For instance, the sequence
$p_3,p_2,\land,\neg,p_1,\lor$
represents the formula $p_1\lor\neg(p_2\land p_3)$ that can be
reconstructed as well as evaluated using a stack, e.g., for any data
point. The encoding deploys a parameter \code{l} for the maximum
length of the sequence/explanation. The dots ``\ldots'' in the first
cell (see Fig.~\ref{fig:layers}) illustrate the fact that the sequence
may start by unused symbol positions whenever a shorter (optimal)
explanation is feasible.  We refer the reader to \cite{jelia23} for
further details of the encoding, but recall that the objective
function maximizes the accuracy \eqref{eq:accuracy} as the primary
criterion and minimizes the length of the formula as the secondary
criterion. In this way, the parameter \code{l} acts just as an upper
bound and the user need not provide the exact length of explanations
in advance.

It should also be emphasized that the search implemented by a conflict-driven learning solver (such as \emph{Clingo-dl)} is far from exhaustive brute-force search. The search space is limited by several factors such as attribute values admitted by the underlying data set, the value of the objective function, as well as conflicts learned during the search. In addition, the encoding of \cite{jelia23} deploys syntactic symmetry constraints to reduce the hypothesis space.

\subsection{Dynamic Booleanization of attributes}
\label{sec:booleanization}

The values of Boolean attributes and one-hot encoded categorical
attributes can be forwarded unchanged through the \code{val/3}
predicate. The dynamic approach is applied to other numeric attributes
that are made available in terms of the \code{rd(D,I,V)} predicate with
arguments
\code{D} for a data point,
\code{I} for the index of an attribute, and
\code{V} for an integer value,
hence assuming some mapping from actual values to the integer domain
(see Section \ref{sec:raw-data}). Then, given an attribute \code{a(I)}
indexed by \code{I}, its minimum and maximum values can be determined
and recorded as respective facts using predicates
\code{minval/2} and \code{maxval/2}.

In the dynamic approach based on a single pivot, the first two rules
in Figure~\ref{fig:layers} choose a value \code{p(I)} in this range.\footnote{Note that the constant bounds $l\leq x\leq u$ for a variable
$x$ can be expressed by $z-x\leq -l$ and $x-z\leq u$ in the normal form,
using an auxiliary variable $z$ standing for ``zero''.
\label{fn:missing-zero}}
The latter two rules compare the actual value \code{V} at a data point
\code{D} against \code{p(I)} and Booleanize \code{V} as \code{0}, if
\code{V} is smaller than \code{p(I)}, and as \code{1}, otherwise. Although
difference atoms are exploited in a corner case
(cf.~footnote~\ref{fn:missing-zero}),
the benefit is that there is no need to sort the values of the
attribute \code{a(I)} beforehand: the Booleanized value can be decided
based on the pivot \code{p(I)}. It should be emphasized that the
choice of pivots is completely dynamic: any values are fine as long
as the required Booleanization of the data can be realized, able to
justify the current hypothesis in terms of accuracy and length---in
harmony with the idea of SMT. 

The generalization for the interval-based Booleanization is obtained
by introducing two pivots, namely \code{l(I)} and \code{u(I)}, instead
of \code{p(I)} for an attribute \code{a(I)}.
The rules in Figure \ref{fig:layers} must be refined
accordingly. While the minimum value of an attribute \code{a(I)} is
the least value for \code{l(I)}, its maximum value gives the greatest
value for \code{u(I)}. Moreover, the difference constraint
\code{\&diff\{u(I) - l(I)\} >= 0}
is needed to make \code{l(I)} smaller than or equal
to \code{u(I)}. Based on their selected values, the value \code{V} of
the attribute \code{a(I)} is Booleanized as \code{1} if \code{V} falls
in the range from \code{l(I)} to \code{u(I)} including endpoints
and \code{0}, otherwise.

\subsection{Treating non-integer numeric values}
\label{sec:raw-data}

Given a \emph{raw} data file, e.g., as a comma-separated values (CSV)
file, the predicate \code{rd/3} can be computed procedurally. We leave
the mapping to integers at the user's discretion. For instance,
non-integer values can be multiplied by suitable constants and rounded
off or truncated at the decimal point. For the dynamic approach based
on pivots, it is more important to preserve the order than the
magnitudes when considering the values of a particular attribute.

\section{Experiments}

\begin{figure}[!tb]
\centering
\resizebox{0.55\textwidth}{!}{
\begin{tikzpicture}
    \begin{axis}[
            reverse legend,
            transpose legend,
            xbar,
            bar width=.2cm,
            width=0.7\textwidth,
            height=1.3\textwidth,
            legend style={at={(0,-0.05)},
                anchor=north west},
            legend columns = 3,
            legend image code/.code={
            \draw [#1] (-0.1cm,-0.1cm) rectangle (0.25cm,0.1cm); },
            axis on top,
            yticklabels={BankMarketing,BreastCancer,Diabetes, GermanCredit, HeartDisease, Hepatitis, StudentDropout},
            ytick=data,
            xtick pos=left,
            ytick style={draw=none},
            ytick = {1,2,3,4,5,6,7},
            y dir=reverse,
            y tick label style={anchor=west,xshift=.15cm, yshift=6.3*\pgfkeysvalueof{/pgfplots/major tick length}},
            nodes near coords,
            visualization depends on={(102-\thisrow{accuracy}) \as \offset},
            node near coords style={shift={(axis direction cs:\offset,0)},/pgf/number format/.cd,fixed zerofill, precision=1},
            nodes near coords align={horizontal},
            xmin = 0.1, xmax = 115,
            xlabel={\%},
            x label style={at={(axis description cs:0.98,0)},anchor=north},
            minor xtick = {100},
            grid={minor}
        ]

        \draw[fill=blue!6!white] (100,0) rectangle (115, 8);

        \addplot[dashed, purple!80!black,fill=purple!10!white, ] 
                table[x=accuracy, y=dataset, col sep=&, row sep=crcr] {
                accuracy &   dataset             & deviation \\
96.6       &   2        & 5 \\
78.0       &   4        & 5 \\
                81.6       &   5        & 5 \\
                87.2       &   6           & 5 \\
};

        \addplot[brown!80!black,fill=brown!30!white, error bars/.cd,
                error bar style = {thick, black},
                error mark options={
                rotate=90,
                mark size=2pt,
                line width=0.8pt
                },
                x dir=both,
                x explicit] table[x=accuracy, y=dataset, x error=deviation, col sep=&, row sep=crcr] {
                accuracy &  dataset &               deviation \\
                89.1    &   1   &       1.5 \\
                97.1    &   2    &       1.7 \\
                76.5    &   3 &   7.0 \\
                73.4      &   4    &     4.1 \\
                79.5    &   5    &       6.8 \\
                83.0    &   6       &       11.8 \\
                87.5    &   7  &       1.5 \\
        };

        \addplot[red!80!black,fill=red!30!white, error bars/.cd,
                error bar style = {thick, black},
                error mark options={
                rotate=90,
                mark size=2pt,
                line width=0.8pt
                },
                x dir=both,
                x explicit] table[x=accuracy, y=dataset, x error=deviation, col sep=&, row sep=crcr] {
                accuracy &  dataset &               deviation \\
                89.4    &   1   &       1.2 \\
                97.2    &   2    &       2.4 \\
                77.3    &   3 &   6.6 \\
                73.7    &   4    &       4.7 \\
                79.2   &   5    &       9.3 \\
                82.1    &   6       &    10.7 \\
                86.2    &   7  &       1.7 \\
        };

        \addplot[teal!80!black,fill=teal!30!white, error bars/.cd,
                error bar style = {thick, black},
                error mark options={
                rotate=90,
                mark size=2pt,
                line width=0.8pt
                },
                x dir=both,
                x explicit] table[x=accuracy, y=dataset, x error=deviation, col sep=&, row sep=crcr] {
                accuracy &  dataset &               deviation \\
                88.8    &   1   &       1.3 \\
                95.4    &   2    &       3.2 \\
                74.2    &   3 &   6.0 \\
                69.7      &   4    &     3.7 \\
                75.9    &   5    &       7.6 \\
                83.8    &   6       &       12.1 \\
                78.6    &   7  &       2.2 \\
        };

        \addplot[magenta!80!black,fill=magenta!30!white, error bars/.cd,
                error bar style = {thick, black},
                error mark options={
                rotate=90,
                mark size=2pt,
                line width=0.8pt
                },
                x dir=both,
                x explicit] table[x=accuracy, y=dataset, x error=deviation, col sep=&, row sep=crcr] {
                accuracy &  dataset &               deviation \\
                88.8    &   1   &   1.1     \\
                95.5    &   2    &  3.1      \\
                73.4    &   3 &  7.0 \\
                67.3      &   4    &  2.4  \\
                75.5    &   5    &   9.6    \\
                82.2    &   6       &  11.9      \\
                85.8    &   7  &  1.3  \\
        };
        
        \addplot[blue!80!black,fill=blue!30!white, error bars/.cd,
                error bar style = {thick, black},
                error mark options={
                rotate=90,
                mark size=2pt,
                line width=0.8pt
                },
                x dir=both,
                x explicit] table[x=accuracy, y=dataset, x error=deviation, col sep=&,            row sep=crcr] {
                accuracy &   dataset             & deviation \\
                88.9    &   1   &   0.9     \\
                95.9    &   2    &  3.5      \\
                73.5    &   3 &  6.4 \\
                66.9      &   4    &  4.3  \\
                71.3    &   5    &   8.3    \\
                81.5    &   6       &  11.5      \\
                84.9    &   7  &  1.1  \\
        };
        
        \legend{UCI repo, XGBoost, Random forest, Median FSM~~~~~~, Pivot FSM~~~~~~, Interval FSM~~~~~~}
    \end{axis}
\end{tikzpicture}
}
\caption{The average test accuracies and standard deviations obtained for each dataset with the pivoted FSM, random forests, XGBoost and the median FSM. Where available, we include also accuracies reported as Baseline Model Performance in the UCI repository, although these are not directly comparable due to the unknown technical particularities behind the UCI numbers.}
\label{fig:resultsum1}
\end{figure}
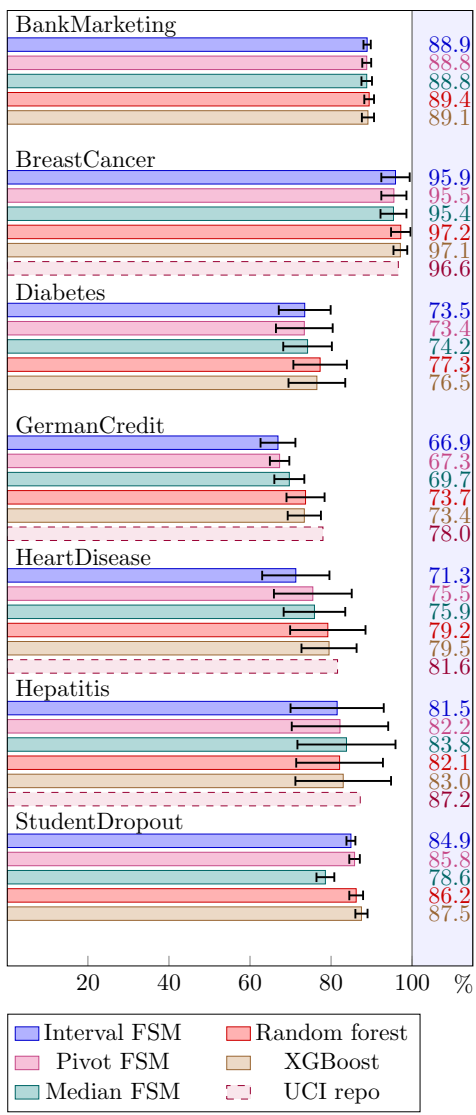

We compare the interval, pivot and median variants of the formula-size method empirically to random forests and XGBoost. The two latter methods are very commonly used state-of-the-art methods for tabular data \cite{swartzarmon22,grinsztajn2022why}. We first present our experimental setup in detail, then discuss the results both in terms of accuracy and the formulas obtained, and finally report the runtimes of the methods.

\subsection{Experimental setup}\label{sec:experimentsetup}

 We test all five methods on seven binary classification tasks with numeric attributes originating from the UCI machine learning repository: BankMarketing, BreastCancer, Diabetes, GermanCredit, HeartDisease, Hepatitis and StudentDropout. The dataset StudentDropout was originally a ternary classification task, but we converted it into a binary one. The dataset Diabetes seems to no longer be available from the UCI repository; it can be downloaded from \url{https://github.com/susanli2016/Machine-Learning-with-Python/blob/master/diabetes.csv}.

The datasets we use contain both categorical and numeric attributes. For categorical attributes we perform a simple one-hot Booleanization. 
For each categorical attribute $X$ and category $a$ of $X$, we create a Boolean attribute $p^X_a$, which is true in the points of category $a$ w.r.t. the attribute~$X$. 

Note that at least the implementations of random forests and XGBoost we use (see the end of this section for details) only support numeric data as input. This means that no feature can be truly categorical as the user is forced to fix an order for the categories by giving them numeric names. We feel that the one-hot encoding described above is closer to a truly categorical attribute than an arbitrary order.

Furthermore, we have experimentally observed that on the datasets that we use random forests and XGBoost seem to obtain better results on one-hot encoded data than on numerically encoded categorical data. Thus we use one-hot encoding as a global preprocessing step for all methods before the experiments.

We move on to numeric attributes. As dynamic discretization is an important feature of our method and the reference methods work directly with numeric attributes, we do not Booleanize numeric attributes beforehand. We do, however, conduct a small preprocessing step to transform floating point attributes into integers for easier compatibility with ASP. We simply multiply the numbers by a suitable power of ten to make them integers. This preprocessing is not done in the case of random forests and XGBoost, which receive the numeric attributes as such.

All missing values have been removed from the datasets. In the case of the Hepatitis dataset, we removed two columns with many missing values. The diabetes dataset does not have missing values, but it has features which take values that are inconsistent with background knowledge (e.g., the dataset contains rows where the value of blood pressure is zero). We removed all rows with such values.

We use ten-fold cross-validation for all methods and all datasets. That is, we randomly split the original data $W$ into ten equal (when possible) parts. For each such 10 percent part, we feed the 90 percent complement into the method as the full training data $W_0$ and set aside the 10 percent part itself as holdout data $W_{\mathrm{hold}}$. After the method has obtained a final classifier using $W_0$, we record the accuracy of that classifier on $W_{\mathrm{hold}}$. We report the average and standard deviation of the accuracy over the ten 90/10 splits given in the beginning.

As the formula size based methods are computationally quite intensive, we employ a timeout for our runs. This raises the possibility that our early stopping trigger is never met before a run times out. In this circumstance, we record the best validation accuracy obtained before timeout and look back at the smallest formula length that achieves the best accuracy. In a large majority of our experiments, the early stopping was triggered before the computation timed out. The exceptions are the dataset StudentDropout and some individual splits scattered among the other datasets.

For all three formula-size methods, the runs were conducted on a Linux cluster featuring Intel Xeon 2.40 GHz CPUs with 8 CPU cores per run, employing a timeout of 72 hours per instance and a memory limit of 64 GB.

For random forests we use the implementation of Scikit-learn. We use nested cross-validation for tuning the hyperparameters for both random forests and XGBoost. As for the formula-size methods, we use a \(70/30\)-split. For hyperparameter optimization, we use Optuna \cite{optuna_paper} with 100 trials. The hyperparameter spaces used for random forests and XGBoost are the same as in \cite{grinsztajn2022why}.
For the readers convenience, we also report the used hyperparameter spaces in the Appendix \ref{appendix:hyperparameter_spaces}. We ran these experiments simply on a laptop.

\begin{table}[tb]
    \begin{center}
    \begin{tblr}{|l|c|c|}
    \hline
    \makecell{Dataset} & \makecell{Example formula} & \makecell{Average \\ length} \\
    \hline
       BM & $p \lor q_{[794,2769]}$  & 2.8\\
    \hline
       BC & $p_{[1,5]} \land q_{[1,6]} \land r_{[1,4]}$ & 5.5\\
    \hline
       D & $p_{[25,60]} \land q_{[128,196]}$ & 4.3\\
    \hline
       GC & $p \lor q_{[4,30]}$ & 3.1\\
    \hline
       HD & $p_{[0.0,2.6]} \land (q \lor (r \land s_{[176,240]}))$ & 2.9\\
    \hline
       H & $p_{[2.9,6.4]}$ & 2.3 \\
    \hline
       SD & $p_{[0,2]} \lor \neg q$ & 4.1\\
    \hline
    \end{tblr}
    \end{center}
    \caption{An example formula obtained by the interval method from the first 90/10 split of each dataset as well as the average length of the ten formulas obtained for all ten 90/10 splits.}
    \label{tab:interval_formulas}
\end{table}

\subsection{Results}\label{sec:results}

\begin{table}[tb]
    \begin{center}
    \begin{tblr}{|l|c|c|}
    \hline
    \makecell{Dataset} & \makecell{Example formula} & \makecell{Average \\ length} \\
    \hline
       BM & \((p \lor q_{\geq 767}) \land \neg r\)  & 4.0 \\
    \hline
       BC & \(\neg (p_{\geq 6} \lor q_{\geq 7} \lor r_{\geq 5})\) & 6.5 \\
    \hline
       D & \(q_{\geq 158} \lor (p_{\geq 29} \land r_{\geq 27.6} \land s_{\geq 109})\) & 2.9 \\
    \hline
       GC & \(\neg r_{\geq 10876}\)  & 3.0 \\
    \hline
       HD & \(\neg t \lor (q \land r)\)  & 4.4 \\
    \hline
       H & \(q\) & 2.8 \\
    \hline
       SD & \(\neg q \land (p_{\geq 2} \lor \neg r_{\geq 1})\) & 6.9 \\
    \hline
    \end{tblr}
    \end{center}
    \caption{An example formula obtained by the pivot method from the first 90/10 split of each dataset as well as the average length of the ten formulas obtained for all ten 90/10 splits.}
    \label{tab:pivot_formulas}
\end{table}

\begin{table}[tb]
    \begin{center}
    \begin{tblr}{|l|c|c|}
    \hline
    \makecell{Dataset} & \makecell{Example formula} & \makecell{Average \\ length} \\
    \hline
       BM & \(q_{m} \land (s \lor p)\) & 6.6 \\
    \hline
       BC & \(\neg ((q_{m} \land s_{m}) \lor (p_{m} \land t_{m} \land  u_{m}))\) & 8.8  \\
    \hline
       D & \(s_{m} \land p_{m} \land (t_{m} \lor q_{m})\) & 4.2 \\
    \hline
       GC & \(s \lor \neg(t \land u)\) & 5.5 \\
    \hline
       HD & \(q\) & 4.2 \\
    \hline
       H & \(q\) & 2.5 \\
    \hline
       SD & \(\neg (p_m \lor (q \land (s \lor t_m)))\) & 7.1 \\
    \hline
    \end{tblr}
    \end{center}
    \caption{An example formula obtained by the median method from the first 90/10 split of each dataset as well as the average length of the ten formulas obtained for all ten 90/10 splits. Propositions that were obtained from medians of numeric attributes are marked with the subscript $m$.}
    \label{tab:median_formulas}
\end{table}

The results of our experiments are summarized in Figure \ref{fig:resultsum1} and Tables \ref{tab:interval_formulas}, \ref{tab:pivot_formulas} and \ref{tab:median_formulas}. In Figure \ref{fig:resultsum1} we report the average accuracy and standard deviation of each method over ten-fold cross-validation. We can see that the formula size methods give accuracies comparable to those of random forests and XGBoost for every dataset. The best performance was obtained in the BankMarketing dataset, where the difference is less than 1 percentage point, while the largest gap of 4 percentage points is found in the GermanCredit dataset. 

To measure interpretability, we report the average length of the final formula for each dataset in Table \ref{tab:interval_formulas} for the interval method, in Table \ref{tab:pivot_formulas} for the pivot method and in Table \ref{tab:median_formulas} for the median method, along with an example formula for each dataset. The example formulas were taken from the first 90/10 split of each ten-fold cross-validation. We see that the formulas we obtain are typically very short with all three methods with the largest average length being 8.8 for the median method on the BreastCancer dataset. Keeping in mind that formula length includes all connectives as well as proposition symbols, this means that all formulas we obtain are easily human interpretable.

Comparing the three formula size methods, we can observe from Tables \ref{tab:interval_formulas}, \ref{tab:pivot_formulas} and \ref{tab:median_formulas} an antitone relationship between the expressive power of a method and the average lengths of formulas produced by it: the more expressive the method is, the shorter the formulas it produces. This makes sense, as the expressive power of a method is related to how quickly the early stopping criteria is met. Furthermore, the more expressive methods are also computationally more intensive, so the timeouts can occur at smaller formula lengths, again shortening the formulas.

From Figure \ref{fig:resultsum1} we can also observe that on these datasets, the greater expressive power of a method does not always translate to better accuracy on the holdout data. 
Indeed, the median method, which has the lowest expressive power, obtains on several datasets accuracies that are similar to or slightly better than the ones obtained via the alternative formula size based methods. 
With no early stopping or cross-validation, the more expressive methods would clearly give a better accuracy on the training data. 
Therefore the explanation for the lower accuracies of the more expressive methods must relate to these testing measures.

When utilizing the same formula length, the interval and pivot methods have a higher susceptibility to overfitting to the training data in contrast to the median method. Although the higher expressive power of the interval and pivot methods may result in shorter formulas and hence improved interpretability, it also heightens the likelihood of generating formulas excessively tailored for the training data, thus lowering holdout accuracy.

Another possible phenomenon is related to the fact that the final classifier is formed using the union $W$ of the training data and the validation data. When forming this classifier, the interval $[r,r']$ or the pivot value $r$ is optimized again, possibly leading to overfitting to $W$ at this final stage. Possible ways to address these issues in the future would be to consider different early stopping criteria or to migrate more information from the validated classifier to the final one.

For the StudentDropout dataset, the interval and pivot methods have obtained significantly better accuracies than the median method. This is because in this dataset, the ability of the more expressive methods to choose a cutoff point other than the median is very useful. The attribute $p$, that occurs in the example formulas of all methods in Tables \ref{tab:interval_formulas}, \ref{tab:pivot_formulas} and \ref{tab:median_formulas} for the StudentDropout dataset, corresponds to the number of approved curricular units a student has completed in the 2nd semester. For this attribute, the pivot and interval methods choose a pivot point of 2, whereas the median of this attribute is 5. Although StudentDropout is the only dataset among those we studied where this behavior is clearly seen, it is nevertheless clear that for many real-world datasets the ability to use dynamic predicates is essential for good accuracy.

\subsection{Runtimes}\label{sec:runtimes}

The average training time of each formula-size method is reported in Table \ref{tab:runtimes}. The runtimes, while considerably large, are still very feasible to run on powerful computation setups that are available today. We do not report the runtimes of random forests and XGBoost as they were all run on a laptop and took less than five minutes. 

We see that in the majority of cases, the more computationally intensive method has a higher runtime. However, there are also cases, where this relationship is reversed. At least one example of this can be seen for any two of the three methods; we use the pivot and median methods as an example here. For these two methods, the reversed relationship occurs on the BankMarketing dataset. There are at least two possible reasons for this.
Firstly, the more expressive pivot method could trigger early stopping at a lower formula length, thus saving runtime compared to the median method.
Secondly, due to its computational intensity, the pivot method can reach timeout sooner, while the median method can keep performing runs for increasing formula lengths and accumulate more runtime.

\begin{table}[tb]  
\begin{center}
    \vspace{10pt}
    \begin{tblr}{|l|r|r|r|}
        \hline 
        \makecell{Dataset} & \makecell{Interval \\ FSM} & \makecell{Pivot~ \\ FSM~~} & \makecell{Median \\ FSM}  \\
        \hline
        BM & 281120 & 98403 & 249263\\
        \hline
        BC & 95868 & 4089 & 2193 \\
        \hline
        D & 190758 & 61903 & 19\\
        \hline
        GC & 103844 & 198504 & 146400\\
        \hline
        HD & 120394 & 3817 & 139\\
        \hline
        H & 26722 & 119 & 47\\
        \hline
        SD & 375093 & 480024 & 260895\\
        \hline
    \end{tblr}
    \end{center}
    \caption{The average training time in seconds per split of each formula-size method for each dataset.}
    \label{tab:runtimes}
\end{table}

\section{Conclusion}
\label{sec:conclusions}

In this work, we introduce a method that extracts immediately human interpretable classifiers from tabular data using dynamic discretizations that are determined on the fly while optimizing the accuracy and the length of the classifier. Based on comparing three variants of the method against state-of-the art methods for classifying tabular data, viz.~random forests and XGBoost, we conclude that explanations obtained by our dynamic methods considered are competitively accurate while much easier to interpret.

As regards future work, the layered architecture described in Section \ref{sec:implementation} opens up several avenues for further research. When it comes to the syntactic form of explanations, it is interesting to consider further Boolean connectives and specific normal forms for the classifiers. Yet further strategies for the discretization of attributes can be obtained as refinements of the pivot/interval-based approaches of this work.

\medskip

\medskip

\noindent
\textbf{Acknowledgments.}\ \ \ 
Tomi Janhunen, Antti Kuusisto, Masood Feyzbakhsh Rankooh and Miikka Vilander were supported by the Academy of Finland consortium project \emph{Explaining AI via Logic} (XAILOG), grant numbers 345633 (Janhunen) and 345612 (Kuusisto). Antti Kuusisto and Miikka Vilander were also supported by the Academy of Finland project \emph{Theory of computational logics}, grant numbers 324435, 328987 (to December 2021) and 352419, 352420 (from January 2022). The author names of this article have been ordered on the basis of alphabetical order. 

\bibliographystyle{plain}
\bibliography{kr-sample,local}

\appendix

\section{Appendix}\label{appendix:hyperparameter_spaces}

Here we report the hyperparameter spaces for random forest and XGBoost. Hyperparameters not listed were kept at their default values.

\smallskip

\noindent\textbf{Random forest} \\

\vspace{-10pt}

\noindent  - \texttt{max\_depth}: None, 2, 3, 4

\noindent  - \texttt{n\_estimators}: Integer sampled from \([9.5,3000.5]\) using log domain

\noindent  - \texttt{criterion}: gini, entropy

\noindent  - \texttt{max\_features}: sqrt, log2, None, 0.1, 0.2, 0.3, 0.4, 0.5, 0.6, 0.7, 0.8, 0.9

\noindent  - \texttt{min\_samples\_split}: 2, 3

\noindent  - \texttt{min\_samples\_leaf}: Integer sampled from \([1.5,50.5]\)

\noindent  - \texttt{bootstrap}: True, False

\noindent  - \texttt{min\_impurity\_decrease}: 0.0, 0.01, 0.02, 0.05

\medskip

\noindent\textbf{XGBoost} \\

\vspace{-10pt}

\noindent- \texttt{max\_depth}: Integer sampled from [1,11]

\noindent - \texttt{n\_estimators}: Integer sampled from [100, 5900] with step-size being 200

\noindent - \texttt{min\_child\_weight}: Float sampled from [1.0, 100.0] using log domain

\noindent - \texttt{subsample}: Float sampled from [0.5, 1.0]

\noindent - \texttt{learning\_rate}: Float sampled from [1e-5, 0.7] using log domain

\noindent - \texttt{colsample\_bylevel}: Float sampled from [0.5, 1.0]

\noindent - \texttt{gamma}: Float sampled from [1e-8, 7.0] using log domain

\noindent - \texttt{reg\_lambda}: Float sampled from [1.0, 4.0] using log domain

\noindent - \texttt{reg\_alpha}: Float sampled from [1e-8, 100.0] using log domain

\end{document}